# DIGITAL ELEVATION MODEL CORRECTION IN URBAN AREAS USING EXTREME GRADIENT BOOSTING, LAND COVER AND TERRAIN PARAMETERS

Chukwuma Okolie[1,2,3,*], Jon Mills[3], Adedayo Adeleke[4], and Julian Smit[5]

[1]Division of Geomatics, University of Cape Town, South Africa
[2]Department of Surveying & Geoinformatics, University of Lagos, Nigeria; cokolie@unilag.edu.ng, cjohnokolie@gmail.com
[3]School of Engineering, Newcastle University, United Kingdom; jon.mills@newcastle.ac.uk
[4]Department of Geography, Geoinformatics and Meteorology, University of Pretoria, South Africa; adedayo.adeleke@up.ac.za
[5]AfriMap Geo-Information Services, Cape Town, South Africa; drjlsmit@gmail.com

**ABSTRACT:**

The accuracy of digital elevation models (DEMs) in urban areas is influenced by numerous factors including land cover and terrain irregularities. Moreover, building artifacts in global DEMs cause artificial blocking of surface flow pathways. This compromises their quality and adequacy for hydrological and environmental modelling in urban landscapes where precise and accurate terrain information is needed. In this study, the extreme gradient boosting (XGBoost) ensemble algorithm is adopted for enhancing the accuracy of two medium-resolution 30-metre DEMs over Cape Town, South Africa: Copernicus GLO-30 and ALOS World 3D (AW3D). XGBoost is a scalable, portable and versatile gradient boosting library that can solve many environmental modelling problems. The training datasets are comprised of eleven predictor variables including elevation, urban footprints, slope, aspect, surface roughness, topographic position index, terrain ruggedness index, terrain surface texture, vector roughness measure, forest cover and bare ground cover. The target variable (elevation error) was calculated with respect to highly accurate airborne LiDAR. After training and testing, the model was applied for correcting the DEMs at two implementation sites. The correction achieved significant accuracy gains which are competitive with other proposed methods. The root mean square error (RMSE) of Copernicus DEM improved by 46 – 53% while the RMSE of AW3D DEM improved by 72 - 73%. These results showcase the potential of gradient boosted trees for enhancing the quality of DEMs, and for improved hydrological modelling in urban catchments.

## 1. INTRODUCTION

For quantitative assessments of environmental processes and hazards in urban areas, one of the critical requirements are reliable digital elevation models (DEMs). Important urban applications of DEMs include change detection, urban monitoring (Sirmacek et al., 2010), site selection and suitability analysis, and flood simulation and modelling. However, DEMs are known to suffer accuracy defects in urban areas due to sensor distortions, source data attributes, and errors inherent in the DEM production methods. Moreover, building artifacts in global DEMs cause artificial blocking of surface flow pathways in hydrological modelling (Liu et al., 2021). These errors compromise their quality and adequacy for hydrological and environmental applications (e.g., flood and watershed modelling) where precise and accurate terrain information is needed. High/very high resolution DEMs e.g., from airborne light detection and ranging (LiDAR) are often prohibitively expensive at the city scale. Several space-borne LiDAR missions have been launched in recent years (e.g., ICESat, ICESat-2, GEDI), but do not provide wall-to-wall elevation coverage. Consequently, medium-resolution synthetic aperture radar (SAR) or photogrammetric global DEMs are a viable option, especially in data-sparse regions.

Several methods including machine learning have been proposed to improve DEM accuracy in urban areas (e.g., Liu et al., 2021; Olajubu et al., 2021; Xu et al., 2021; Hawker et al., 2022). For example, Kim et al. (2020) integrated Sentinel-2 multispectral imagery with an artificial neural network (ANN) for improving the accuracy of 30 m shuttle radar topography mission (SRTM) DEM in dense urban cities. Similarly, Liu et al. (2021) adopted the random forest model for the correction of building artifacts in the MERIT DEM using publicly available datasets such as global population density, satellite night-time lights and OpenStreetMap buildings. Hawker et al. (2022) applied random forest for the removal of forests and building offsets from the 30 m Copernicus DEM to produce a globally corrected DEM product referred to as FABDEM.

Tree-based ensemble machine learning algorithms have received significant attention as one of the most reliable and broadly applicable classes of machine learning approaches. Decision trees provide a straightforward interpretation and understanding of the relationships between objects at different levels of detail (Miao et al., 2012). It also has a high tolerance of multicollinearity (Climent et al 2019; Han et al., 2019; Pham and Ho, 2021). Despite the appeal of tree-based models, the remote sensing community is yet to harness their full potential for DEM correction and/or enhancement. Several important terrain conditioning factors and parameters were not considered in previous studies, and there are still some unknowns regarding the interdependence between terrain parameters and their specific contributions to machine learning predictions, when applied to DEM correction.

Among the tree-based machine learning methods in use, gradient tree boosting has excelled in a myriad of applications, and has been shown to give state-of-the-art results (e.g., Samat et al., 2020; Bentéjac et al., 2021; Łoś et al., 2021). In this study, the extreme gradient boosting (XGBoost) algorithm is adopted. XGBoost is a scalable end-to-end tree boosting system that is commonly used by data scientists, provides state-of-the-art results on many problems, and has excelled in numerous data mining and machine learning challenges (Chen and Guestrin, 2016). The predominant factor behind the success of XGBoost is its scalability in multiple scenarios. This study aims to integrate XGBoost with terrain and land cover parameters for urban correction of Copernicus and ALOS World 3D DEMs in a section of Cape Town, South Africa.

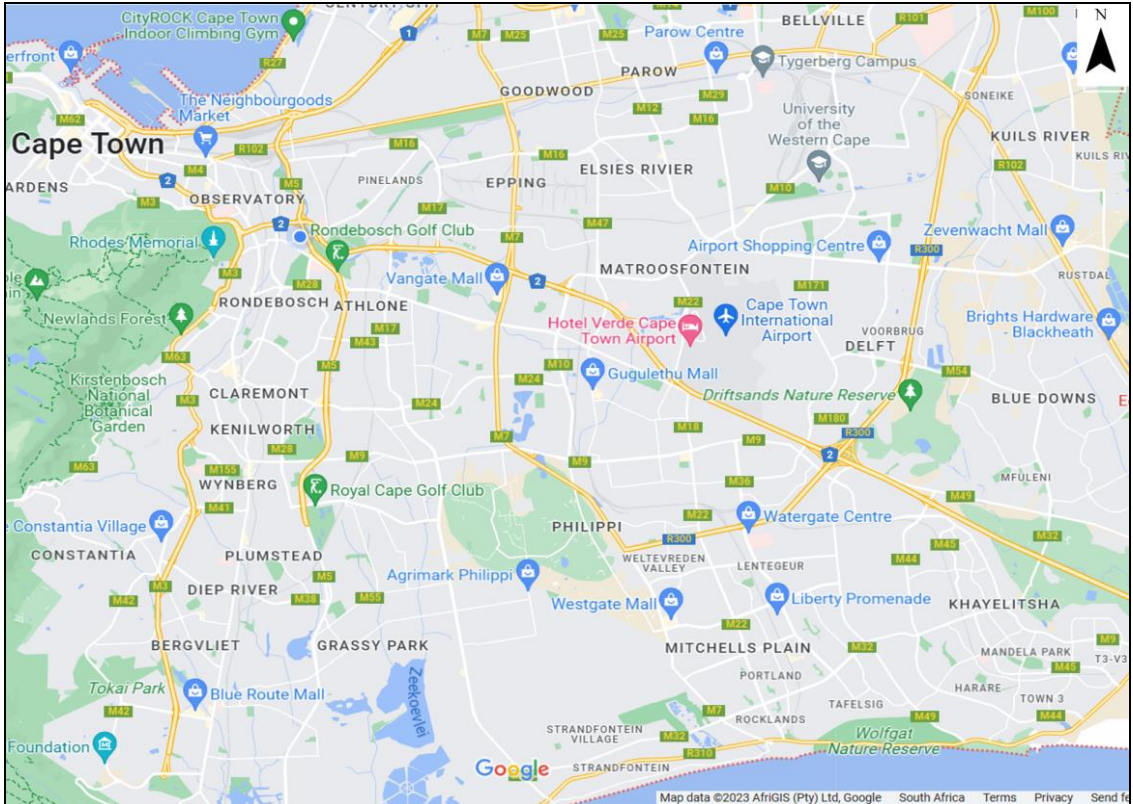

**Figure 1.** Map view of the urban setting within the Cape Town metropolitan municipality, from which data for training, validation, testing and implementation were selected

## 2. METHODOLOGY

### 2.1 Study Area

The City of Cape Town in South Africa is a large urban area with an intense movement of people, goods and services, high population density, extensive developments, industrial areas and multiple business districts (Western Cape Government, 2022).

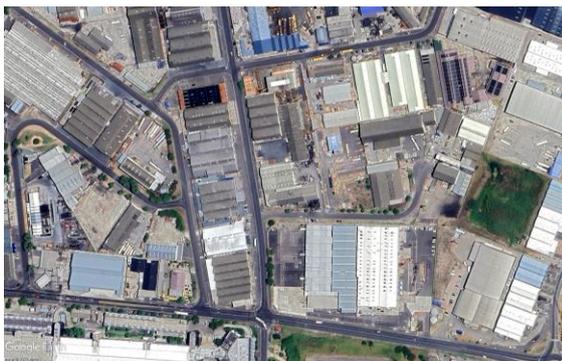

**Figure 2.** Satellite image close-up view showing buildings in the urban area

Cape Town is a complex and diverse city, and the second-largest city in South Africa, with a population of over four million people (Smit, 2020). The city includes the Cape Metropolitan Council, Cape Town Central Business District, the South Peninsula and other areas. The sites for this study are selected from urban districts in Cape Town (Figures 1 – 2).

### 2.2 Datasets

**2.2.1 Digital elevation models**: Two 30 m global digital elevation models (DEMs) are considered in this study, the Copernicus GLO-30 DEM and ALOS World 3D DEM (AW3D30). Copernicus GLO-30 is derived from the WorldDEM data. The WorldDEM data product is based on the radar satellite data which was acquired during the TanDEM-X Mission (ESA, 2020). AW3D30 was released by the Japan Aerospace Exploration Agency (JAXA). It was generated from the earlier ALOS DEM which was produced at a spatial resolution of 5 m with an accuracy of 5 m (standard deviation) (JAXA 2017).

The most recent versions (Copernicus GLO-30 and AW3D v3.2) were adopted in this study. The City of Cape Town (CoCT) airborne LiDAR-derived DEM is used as the reference dataset. It was acquired from the Information and Knowledge Management Department of the City of Cape Town. The height accuracy of the point cloud used for generating the DEM is 0.15 m. The LiDAR DEM is spatially referenced to the Hartebeesthoek94 horizontal co-ordinate system and vertically referenced to the South Africa (SA) Land Levelling Datum (SAGEOID2010).

**2.2.2 Global urban footprint:** The Global Urban Footprint (GUF) is a human settlement layer that was created from the global synthetic aperture radar (SAR) dataset that was acquired during the TanDEM-X (TDM) mission. The methodology behind the classification approach used for deriving the GUF is presented in Esch et al. (2010, 2013). The high-resolution 0.4 arc-second (~12 m) GUF 2012 product was adopted for this study.

**2.2.3 Terrain parameters and land cover data:** To characterise the influence of the terrain on the elevation error, the following additional input variables which are known influencers of DEM error were selected: elevation, slope, aspect, surface roughness, topographic position index (TPI), terrain ruggedness index (TRI), terrain surface texture (TST), vector ruggedness measure (VRM), percentage forest canopy and percentage bare ground cover. The elevation errors or differences (ΔH) between the DEMs and reference LIDAR were calculated as follows:

$$\Delta H = H_{Global\ DEM} - H_{RefDEM} \tag{1}$$

Where,

$H_{RefDEM}$ = elevations from LiDAR DEM.

$H_{Global\ DEM}$ = individual elevations from the global DEMs (i.e., Copernicus and AW3D)

**2.3 Extreme Gradient Boosting**

Extreme Gradient Boosting (XGBoost) is a scalable end-to-end tree boosting system that sequentially builds short and simple decision trees as each tree attempts to improve the performance of the previous tree (Chen and Guestrin, 2016; Safaei et al., 2022). It parallelizes each tree's training and speeds up the training process (Safaei et al., 2022).

Gradient boosting combines weak learners into strong learners in an iterative fashion (Friedman, 2001; Hastie et al., 2009; Deng et al., 2019; Can et al., 2021). During each iteration, the residual is used for correcting the previous predictor in order to optimise the loss function. The core algorithm of XGBoost is its optimisation of the objective function, $O(\theta)$, which consists of the training loss and regularisation (Deng et al., 2019; Can et al., 2021):

$$O(\theta) = L(\theta) + \Omega(\theta) \tag{2}$$

Where:
$L$ – training loss function
$\Omega$ – regularisation term

The training loss is used to evaluate the performance of the model on training samples, while the regularisation term controls the model complexity. The regularisation term for a decision tree can be defined as (Deng et al., 2019):

$$\Omega(f) = \gamma T + \frac{1}{2}\lambda \sum_{j=1}^{T} w_j^2 \tag{3}$$

Where,
$T$ - the number of leaves in a decision tree
$w$ - the vector of scores on leaves
$\gamma$ - the complexity of each leaf
$\lambda$ - a parameter to scale the penalty

The objective function for calculating the structure score of XGBoost is derived as (Deng et al., 2019):

$$O = \sum_{j=1}^{T}\left[g_j w_j + \frac{1}{2}(h_j + \lambda)w_j^2\right] + \gamma T \tag{4}$$

where $w_j$ are independent with each other. The form $g_j w_j + \frac{1}{2}(h_j + \lambda)w_j^2$ is quadratic and the best $w_j$ for a given structure is $q(x)$.

**2.4 Data Preparation**

To harmonise the horizontal datums, the global DEMs were transformed from the geographic to the Universal Transverse Mercator (UTM) projection in WGS84. Similarly, the LiDAR DEM was transformed from Hartebeesthoek94 to UTM WGS84. The vertical datum of the global DEMs and airborne LiDAR were harmonised to EGM2008.

| DEM | Training, validation and testing | Implementation site A | Implementation site B |
|---|---|---|---|
| Copernicus | 573377 | 23041 | 22988 |
| AW3D | 572374 | 23041 | 22988 |

**Table 1**. Distribution of model training, validation, test and implementation sites data

To derive the elevation errors (ΔH), the LiDAR elevations were subtracted from the corresponding elevations of the global DEMs at specific points. The elevation values, along with the values of the elevation error and terrain/land cover parameters were extracted from the rasters to csv files, and sorted. This resulted in the final set of points used for model training, validation and testing split into 80% for training and validation, and 20% for testing. After training and testing, the models were independently evaluated at external sites referred to as model implementation sites A and B respectively. The model implementation provides an opportunity for evaluation of the prediction capability and accuracy of the trained models. Table 1 shows the data distribution.

**2.5 Model Implementation**

The model was trained using the elevation, urban footprints, slope, aspect, surface roughness, topographic position index, terrain ruggedness index, terrain surface texture, vector ruggedness measure, percentage forest canopy and percentage bare ground cover as input parameters, and the elevation error as the target variable (or predictand).

The training was carried out using the default hyperparameters and tuned hyperparameters. For the hyperparameter tuning, Bayesian optimisation was adopted. The theoretical background of Bayesian optimisation is already well documented in the extant literature. Essentially, it "provides a principled technique based on Bayes Theorem to direct a search of a global optimization problem that is efficient and effective" (Brownlee, 2019). The explanations of the XGBoost hyperparameters are provided in the XGBoost library documentation (XGBoost, 2022). Summarily, the adopted hyperparameters and the search space are shown in Table 2. After training and testing, the models were saved, loaded and implemented for predicting the height errors at implementation sites (A and B) with similar terrain characteristics. The predicted elevation errors were applied for deriving the corrected DEMs (i.e., $DEM_{Corrected} = DEM_{Original} - \Delta H_{Predicted}$).

## 2.6 Model Performance and Accuracy Assessment

Learning curves are used as a diagnostic tool for evaluating the performance of XGBoost on the training and validation datasets. In the following analysis, the learning curves are reviewed to assess the representativeness of the training and validation datasets, and any possible problems with underfitting or overfitting. Two learning curves are analysed: (i) training curve: to give an idea of how well the model is learning, and (ii) validation curve which gives an idea of how well the model is generalising. The training and test error were compared using

## 3. RESULTS AND DISCUSSION

### 3.1 Model Diagnostics

In the learning curves for both comparisons (default vs. Bayesian; Figure 3), the plot of the training and validation error decreases with successive epochs. In the models with default hyperparameters, the gap between the training and validation curves is minimal within the 100 epochs shown. The Bayesian tuning of XGBoost increased the number of estimators (n_estimators) significantly, thus leading to a higher number of the root mean square error (RMSE) metric. The accuracy of the corrected DEMs was also assessed using the RMSE. The RMSE is very common, and is considered an excellent general-purpose error metric for assessing numerical predictions (Christie and Neill, 2022). The model explainability is addressed using feature importance plots.

| Description of hyperparameter | Name/alias | Search space |
|---|---|---|
| No of trees | n_estimators | (0, 2000) |
| Max depth of tree | max_depth | (1, 10) |
| Learning rate | learning_rate | (0.001, 1) |
| Regularisation parameters | reg_alpha | (0.001, 10) |
|  | reg_lambda | (0.001, 10) |
| Others | subsample | (0.001, 1.0) |
|  | colsample_bytree | (0.001, 1.0) |
|  | min_child_weight | (0.001, 10) |
|  | gamma | (0.001, 10) |

**Table 2.** The final selected hyperparameters for Bayesian optimization of XGBoost

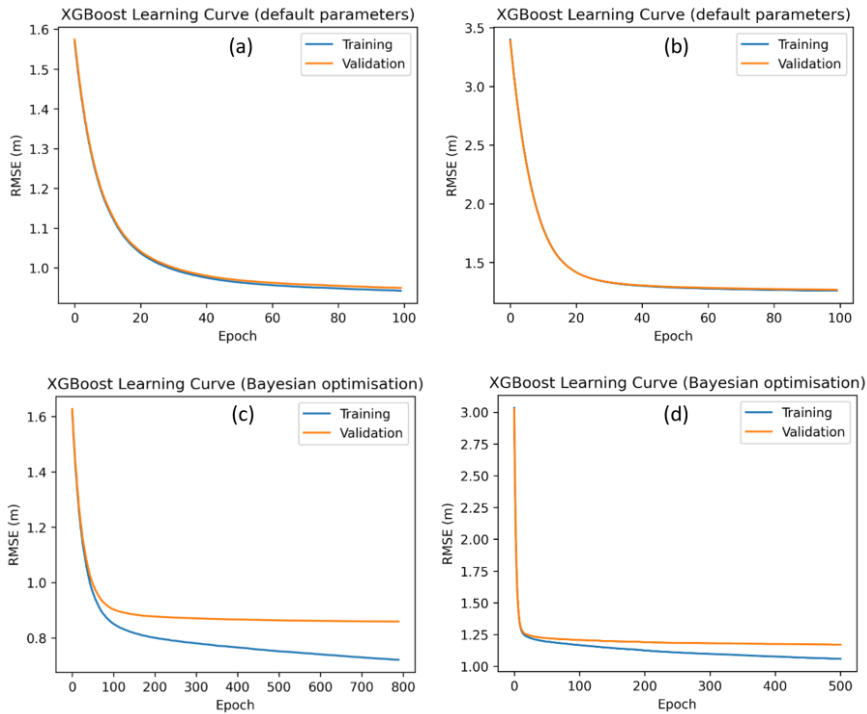

**Figure 3.** Learning curves showing error loss at successive training epochs, (a) Copernicus, default (b) AW3D, default (c) Copernicus, Bayesian-optimised (d) AW3D, Bayesian-optimised

| DEM | RMSE - Default hyperparameters (m) | | RMSE - Bayesian optimisation (m) | |
|---|---|---|---|---|
|  | Training | Test | Training | Test |
| Copernicus | 0.94 | 0.95 | 0.72 | 0.86 |
| AW3D | 1.26 | 1.27 | 1.06 | 1.17 |

**Table 3.** Comparison of training and test RMSEs

training epochs, that ranged from 500 – 800. Even with the Bayesian optimised model validation accuracy plateau-ing early, it is still better than the accuracy of the default model. Early stopping was implemented for both models to stop training at any point where there was no further improvement in the RMSE after 10 rounds. Expectedly, the training errors at both sites are lower than the validation error, because the models are fit on the training data. Training and test RMSEs were calculated and are compared in Table 3.

**3.2 Accuracy of Corrected DEMs**

Figure 4 shows a visual comparison of the original and corrected DEMs at implementation site B. The corrections achieved more realistic terrain representations in both Copernicus and AW3D as shown by the closer consonance of the corrected DEMs to the LiDAR groundtruth. The height error

Other positive changes in the corrected DEMs include:

- Diminution of surface artifacts and discontinuities
- Minimisation of building artifacts
- Reduction in noise effects and high frequency contents.
- Significant improvement in the representation of terrain details.
- Minimisation of elevation error dispersion

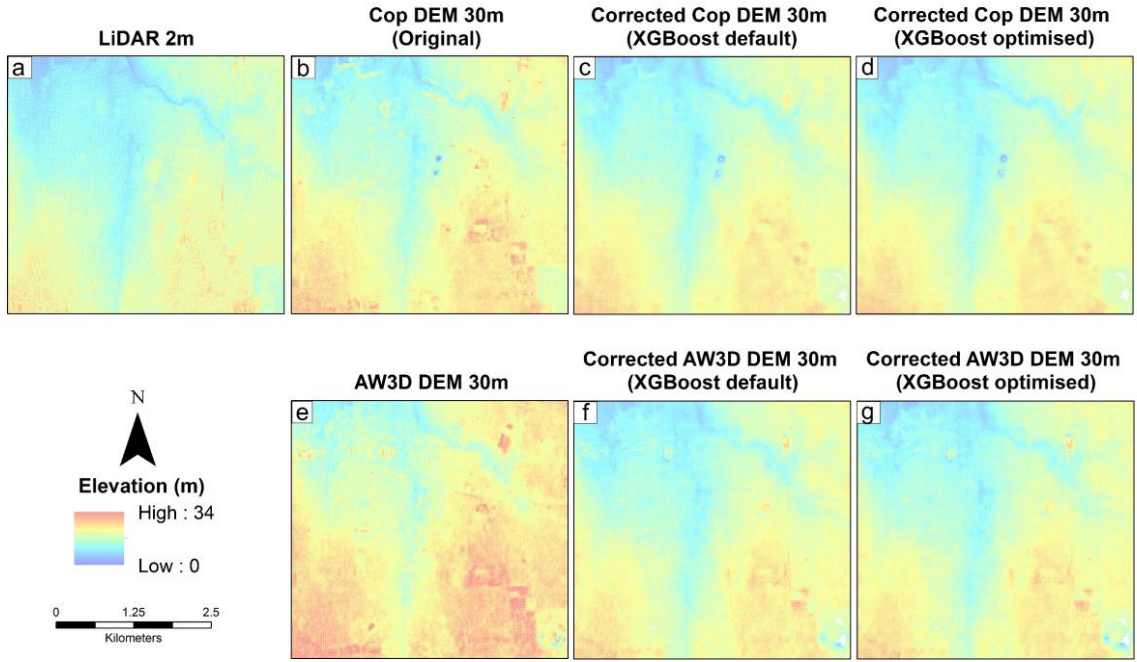

**Figure 4.** Visual comparison of the DEMs, original *vs* corrected at site B, LiDAR (a), Copernicus (b – d) and AW3D (e – g)

| Site | DEM | Original DEM RMSE (m) | Corrected DEM RMSE (m) | |
|---|---|---|---|---|
| | | | Default hyperparameters | Bayesian Optimisation |
| A | Copernicus | 1.859 | **0.877** | 0.888 |
| | AW3D | 4.783 | 1.324 | **1.316** |
| B | Copernicus | 1.470 | 0.788 | **0.771** |
| | AW3D | 3.628 | **0.987** | 1.005 |

**Table 4.** Accuracy of original DEM versus corrected DEM, default hyperparameters *vs* Bayesian optimisation

maps (Figures 5 and 6) show significant reduction in the elevation error, after correction. Notable improvements in accuracy are observed in Copernicus and AW3D DEMs using both the default and optimised models. The optimised model outperformed the default model in some cases. At site A, the correction to Copernicus DEM achieved a reduction in RMSE from 1.859 m to 0.877 m and 0.888 m in the default and optimised models respectively.

At site B, the correction improved the RMSE of Copernicus from 1.470 m to 0.788 m and 0.771 m in the default and optimised models respectively. Similarly, there were improvements in the accuracy of AW3D (RMSE reduction from 4.783 m to 1.324 m and 1.316 m respectively at site A; and reduction from 3.628 m to 0.987 m and 1.005 m at site B, respectively). This represents an improvement factor of 46 – 53% in Copernicus, and 72 – 73% in AW3D.

**3.3 Analysis of Feature Importance**

The topographic position index (TPI) and elevation are the most important features influencing the prediction of elevation error by the default and Bayesian-optimised models, respectively (Figure 7). With the default model, the top three most influential parameters are the TPI, terrain surface texture (TST) and elevation, whereas with the Bayesian-optimised model the top three most influential features are the TST, VRM and elevation.

Generally, the F-scores of the features in the Bayesian-optimised version are orders of magnitude higher than the default mode, suggesting the capability of Bayesian optimisation to exploit more complex interaction between the variables for better predictions. However, it should be noted that the influence or relevance of features in the outcome of machine learning predictions is terrain-dependent.

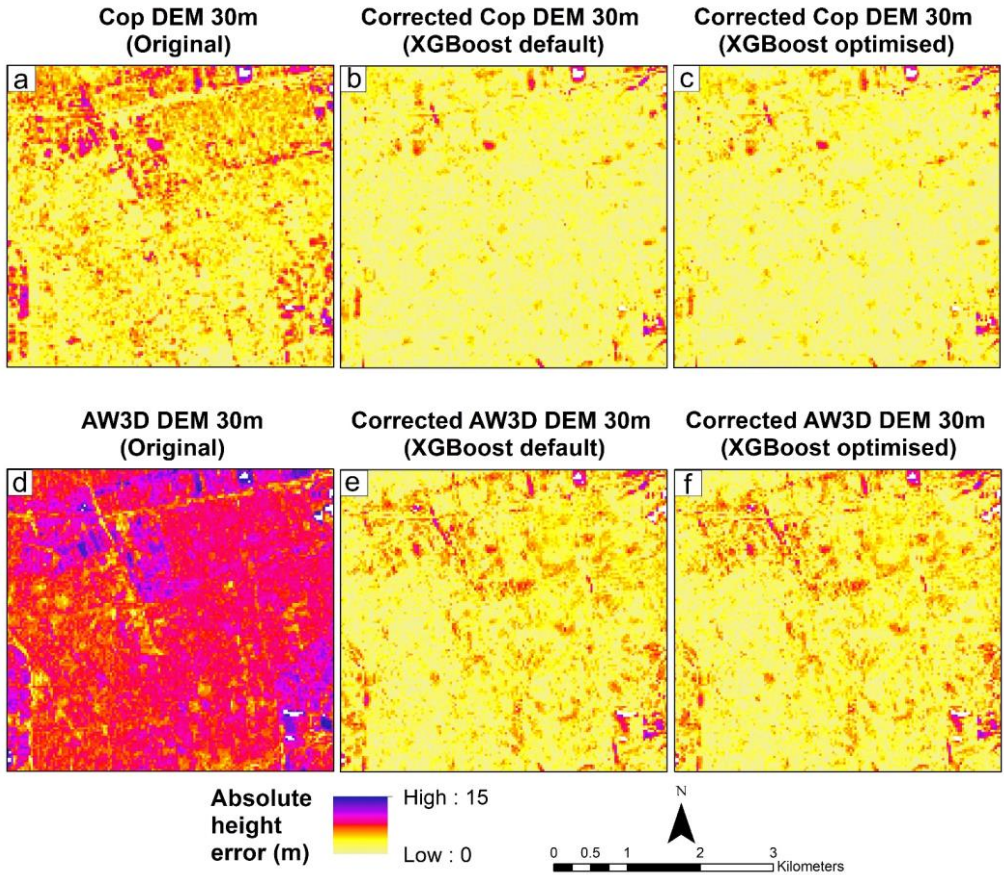

**Figure 5.** Height error maps of original *vs* corrected DEMs at site A, Copernicus (a – c) and AW3D (d – f)

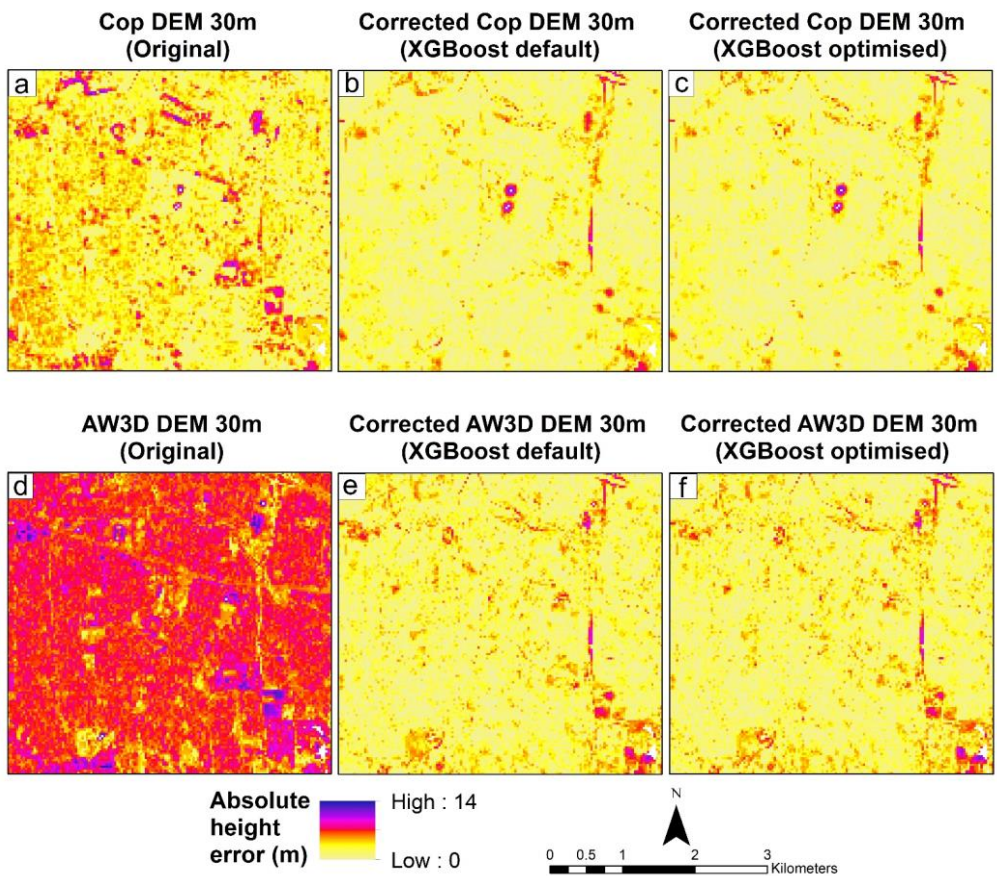

**Figure 6.** Height error maps of original *vs* corrected DEMs at site B, Copernicus (a – c) and AW3D (d – f)

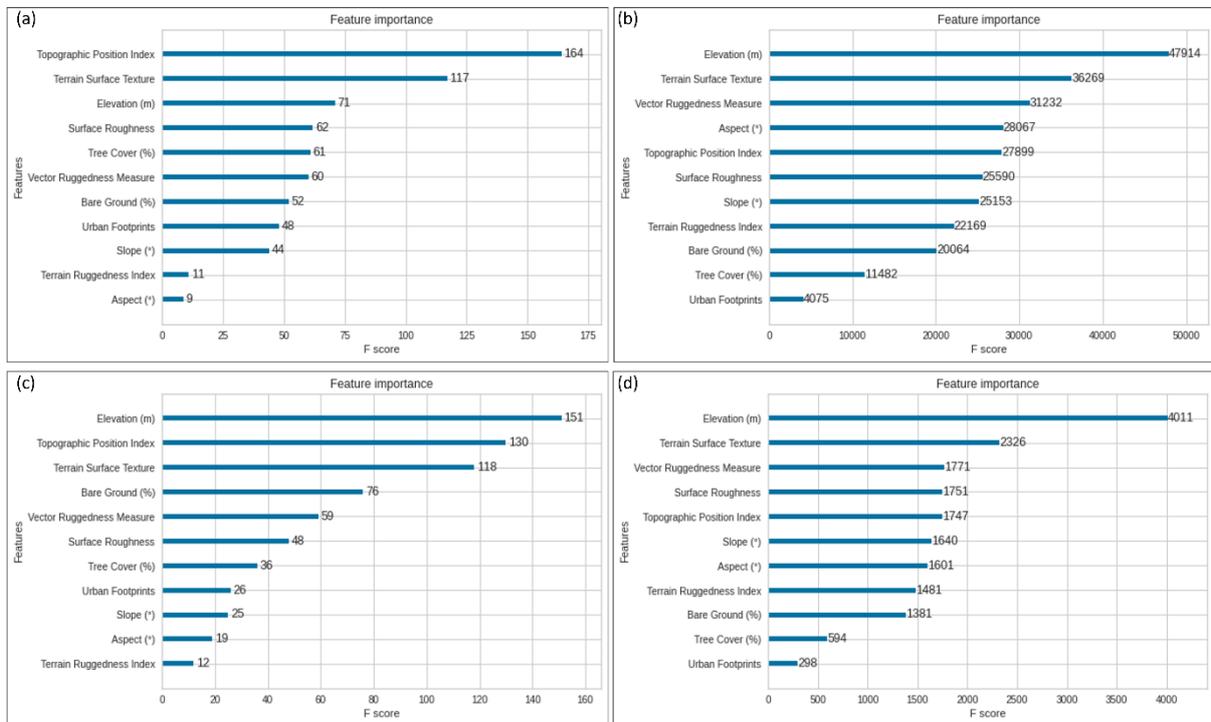

**Figure 7.** Feature importance plots of the input variables for height error prediction, (a) Copernicus, default (b) Copernicus, Bayesian-optimised (c) AW3D – default (d) AW3D, Bayesian-optimised

## 4. CONCLUSION

Generally, the corrected DEMs show several topographic improvements such as smoothing of rough edges, enhanced stream channel conditioning and diminution of coarse/grainy pixels. The elevation error dispersion has also been reduced in the corrected DEMs. The results also show that hyperparameter optimization with Bayesian optimization can yield appreciable gains in accuracy. Thus, tuning the hyperparameters of tree-based models is recommended as a measure to improve the accuracy of predictions. The topographic position index and elevation were the most influential features in the default and optimised model.

The methodology presented in this study is simple, low-cost and easy-to-follow. Moreover, the ensemble framework can learn non-linear and multi-variate spatial patterns in urban environments. The corrections are implemented on a point-by-point basis, in contrast to other techniques that only address the global bias. The introduced methodology based on the integration of XGBoost, land cover and terrain parameters shows great potentials for improved hydrological modelling in urban catchments.

Since machine learning algorithms are likely to be biased to the landscape characteristics fed into them, future research can explore the performance of the proposed approach in different landscapes.

## ACKNOWLEDGEMENTS

Special thanks to the Commonwealth Scholarship Commission UK, and the University of Cape Town Postgraduate Funding Office for funding support for this research. LIDAR data for the City of Cape Town was provided by the Information and Knowledge Management Department, City of Cape Town. Also, thanks to Ikechukwu Maduako, Hossein Bagheri, Tom Komar, Shidong Wang, Maria Peppa, Chima Iheaturu and Ikenna Arungwa for their insightful feedback.